\documentclass[conference]{IEEEtran}
\IEEEoverridecommandlockouts
\usepackage{cite}
\usepackage{amsmath,amssymb,amsfonts}
\usepackage{algorithmic}
\usepackage{graphicx}
\usepackage{textcomp}
\usepackage{xcolor}
\def\BibTeX{{\rm B\kern-.05em{\sc i\kern-.025em b}\kern-.08em
    T\kern-.1667em\lower.7ex\hbox{E}\kern-.125emX}}
\begin{document}

\title{Automated Scoring of Clinical Patient Notes using Advanced NLP and Pseudo Labeling\\

}

\author{\IEEEauthorblockN{Jingyu Xu}
\IEEEauthorblockA{\textit
\textit{Northern Arizona University}\\
Flagstaff, USA \\
jyxu01@outlook.com}
\and
\IEEEauthorblockN{Yifeng Jiang}
\IEEEauthorblockA{\textit
\textit{Boston University}\\
 Boston, USA\\
yjiang8@bu.edu}
\and
\IEEEauthorblockN{Bin Yuan}
\IEEEauthorblockA{
\textit{Trine University}\\
Phoenix, USA \\
binyuan2235@gmail.com}
\and
\IEEEauthorblockN{Shulin Li}
\IEEEauthorblockA{
\textit{Trine University}\\
Phoenix, USA \\
liam.cool666@gmail.com}
\and
\IEEEauthorblockN{Tianbo Song\textsuperscript{*}}
\IEEEauthorblockA{
\textit{Arizona State University}\\
Phoenix, USA \\
tianbosong@aol.com}
}

\maketitle

\begin{abstract}
Clinical patient notes are critical for documenting patient interactions, diagnoses, and treatment plans in medical practice. Ensuring accurate evaluation of these notes is essential for medical education and certification. However, manual evaluation is complex and time-consuming, often resulting in variability and resource-intensive assessments. To tackle these challenges, this research introduces an approach leveraging state-of-the-art Natural Language Processing (NLP) techniques, specifically Masked Language Modeling (MLM) pretraining, and pseudo labeling. Our methodology enhances efficiency and effectiveness, significantly reducing training time without compromising performance. Experimental results showcase improved model performance, indicating a potential transformation in clinical note assessment.
\end{abstract}


\section{Introduction}
Clinical patient notes play a pivotal role in medical practice, serving as a primary means of documenting patient interactions, diagnoses, and treatment plans. Accurate assessment and scoring of these notes are of paramount importance in the fields of medical education and certification, where they are used to evaluate the competence of healthcare professionals and maintain the quality of patient care.

Scoring clinical patient notes is a complex and time-consuming task, demanding a deep understanding of natural language, context interpretation, and medical terminology. Human assessors often face challenges in reviewing these notes, leading to potential interrater variability and resource-intensive evaluations.

To address these challenges, this research presents a comprehensive approach to automate the scoring of clinical patient notes by harnessing advanced Natural Language Processing (NLP) and Named Entity Recognition (NER) techniques. Our study primarily focuses on leveraging the classic Masked Language Modeling (MLM) pretraining approach, which has shown significant success in various NLP tasks. This approach aims to enhance the understanding and analysis of clinical narratives by capturing intricate relationships and nuances within the text.

Our methodology includes several innovative components to improve both the efficiency and effectiveness of automated clinical patient note scoring. These components encompass data preprocessing techniques to ensure optimal data formatting and tokenization for NLP analysis. Furthermore, we introduce the concept of pseudo-label generation, a strategy that extends the training dataset with additional unlabeled examples, enhancing the model's ability to generalize and capture domain-specific patterns. Additionally, we employ training acceleration techniques to expedite model convergence, significantly reducing training time without compromising performance.

The primary contributions of this work are summarized as follows:
\begin{enumerate}
    \item \textbf{MLM Pretraining Application:} We demonstrate the adaptability and effectiveness of the MLM pretraining approach in the specialized domain of clinical patient note scoring.
    \item \textbf{Pseudo-Label Generation:} The introduction of pseudo-label generation extends our training data, resulting in improved model generalization and performance.
    \item \textbf{Training Acceleration:} Our approach significantly reduces the time required for model training, making it more practical and efficient for real-world applications in medical education and certification.
\end{enumerate}

Through rigorous experimentation, our results showcase the superiority of our methodology compared to baseline models, indicating its potential to revolutionize the automated assessment of clinical patient notes. In summary, this research offers an innovative solution to a critical problem in the medical field, combining advanced NLP techniques with practical considerations for efficiency and accuracy, ultimately contributing to the advancement of medical education and certification processes.

\section{Related Work}

The field of Natural Language Processing (NLP) has witnessed significant progress in various applications, including the evaluation and scoring of clinical patient notes. The integration of NLP in healthcare has led to advancements in diagnostic coding, clinical decision support, and patient triage by extracting relevant features from unstructured text. Researchers have leveraged NLP techniques to tackle the inherent ambiguity and complexity of clinical narratives, with a focus on enhancing the accuracy of information extraction\cite{khanbhai2021applying,nuthakki2019natural}.
Clinical Named Entity Recognition (NER) serves as a foundational NLP task in this domain, with tools like cTAKES, CLAMP, and MetaMap being notable for their abilities to identify and categorize terms from patient notes into predefined categories such as medications, procedures, and diseases\cite{peng2020natural,sheikhalishahi2019natural}. These techniques, powered by machine learning algorithms, have played a pivotal role in linking patient notes to relevant medical codes and terminologies, thus supporting the automated scoring of clinical notes\cite{turchioe2022systematic}.

Furthermore, deep models based on the Transformer architecture\cite{zhang2023trep} have demonstrated remarkable capabilities in an increasing number of domains, surpassing the performance of traditional machine learning methods by a significant margin. Hu et al \cite{hu2023m} introduces M-GCN, a novel approach that excels in classifying 3D point clouds. The advent of transfer learning has enabled the fine-tuning of these pre-trained models on domain-specific corpora, resulting in improved performance on specialized tasks\cite{zhang2022natural}.

The application of NLP to clinical notes has proven invaluable in psychiatric evaluations, where linguistic markers can offer insights into patients' cognitive states. Recent studies have demonstrated the utility of NLP in identifying signs of mental illnesses through language patterns, contributing to early detection and treatment of psychiatric conditions\cite{crema2022natural}.

However, despite these advancements, challenges persist in handling the idiosyncrasies of clinical language, including abbreviations, jargon, and non-standard terminologies. This has led to the development of custom NLP systems tailored to the medical domain, such as cTAKES, which have exhibited higher precision in concept extraction compared to general-purpose NLP tools\cite{mustafa2021automated}.

The availability of annotated medical datasets has significantly advanced the field, enabling the training of more accurate and robust models. Datasets like MIMIC-III have been valuable resources for the research community, supporting a wide range of NLP applications in clinical settings\cite{viani2021natural}.

NLP's role in improving the interpretability of Electronic Health Record (EHR) documentation is paramount. Techniques leveraging word embeddings have been particularly effective in developing question-answering systems and summarizing extensive clinical notes, streamlining the review process for healthcare providers\cite{chen2022applications}.

EMRM improves rating predictions in recommendation systems by leveraging diverse multi-source auxiliary reviews, surpassing baseline methods on real-world datasets \cite{wang2021emrm}.
In the context of scoring clinical patient notes, evaluation metrics such as the micro-averaged F1 score have been employed to measure NLP model performance. These metrics assess the precision and recall of identified spans against predefined ground-truth annotations, ensuring the relevance and accuracy of extracted information\cite{sheikhalishahi2019natural}.

Previous studies have laid foundational work in the application of NLP to clinical notes. However, they often struggled to accurately capture the intricate nuances of medical language and faced limitations in handling the diversity of clinical scenarios. In contrast, our work introduces advanced methods such as MLM pretraining and strategic pseudo-labeling tailored for the medical domain. These innovations have significantly enhanced the model's performance, particularly in deciphering complex medical narratives and addressing challenges unmet by conventional approaches. Our methodology sets a new benchmark in the field, demonstrating the potential of tailored NLP techniques to enhance the accuracy and reliability of clinical note evaluation.

\section{Algorithm and Model}
\begin{figure}[h]
    \centering
    \includegraphics[width=0.3\textwidth]{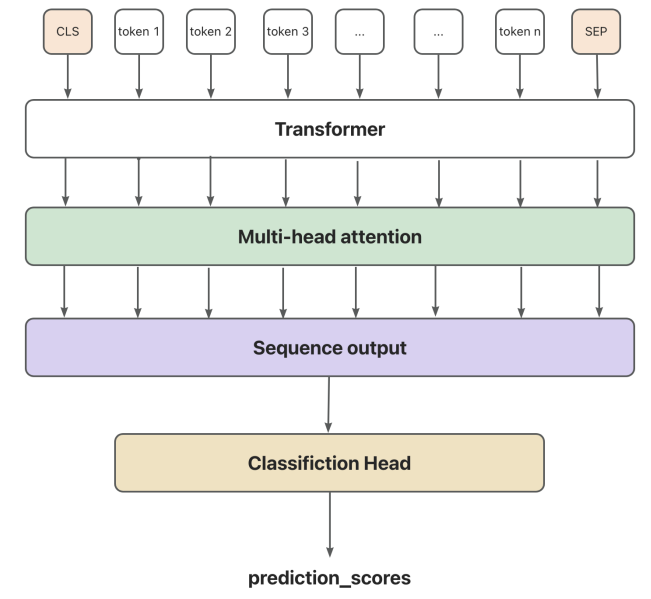}
    \caption{Architecture of DeBERTa}
    \label{fig:DeBERTa}
\end{figure}

DeBERTa is a variant of the BERT architecture that introduces innovative features like Disentangled Attention and Decoding Enhancement. These features enable DeBERTa to better capture semantic relationships and context information in textual data, making it suitable for the task of evaluating clinical patient records. The architecture of DeBERTa is shown in Figure \ref{fig:DeBERTa}.

\subsection{Transformer Architecture}

\begin{figure}[h]
    \centering
    \includegraphics[width=0.3\textwidth]{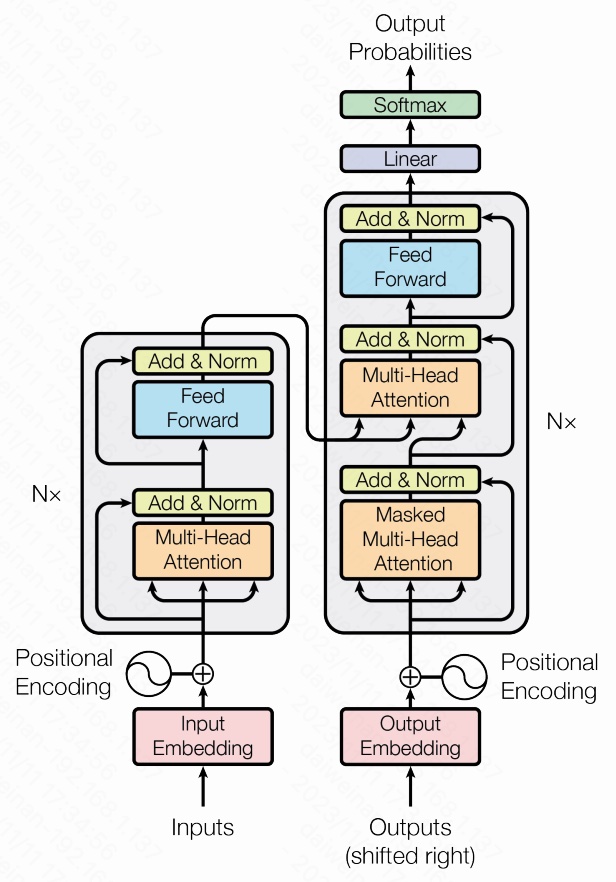}
    \caption{Architecture of Transformer}
    \label{fig:transformer}
\end{figure}

Similar to BERT, DeBERTa is based on the Transformer architecture, as depicted in Figure \ref{fig:transformer}, which consists of encoders and decoders, with BERT representing the encoder part.

\subsubsection{Self-Attention Mechanism}

The key to the Transformer architecture lies in the self-attention mechanism, which allows the model to dynamically assign different attention weights to different positions in the input sequence, capturing dependencies between contexts. The self-attention mechanism is mathematically described as follows:

\begin{equation}
\text{Attention}(Q, K, V) = \text{Softmax}\left(\frac{QK^T}{\sqrt{d_k}}\right)V
\end{equation}

Here, \(Q\), \(K\), and \(V\) represent matrices for queries, keys, and values, and \(d_k\) is the dimension of attention heads. The self-attention mechanism calculates attention weights based on the similarity between queries and keys, enabling the model to focus on relevant information.

\subsubsection{Multi-Head Attention}

Each self-attention layer in Transformer contains multiple attention heads, allowing the model to simultaneously focus on different positions and enhancing its modeling capacity. The multi-head attention mechanism can be mathematically represented as follows:

\begin{equation}
\text{MultiHead}(Q, K, V) = \text{Concat}(\text{head}_1, \text{head}_2, ..., \text{head}_h)W^O
\end{equation}

Where \(\text{head}_i = \text{Attention}(QW_i^Q, KW_i^K, VW_i^V)\), \(W_i^Q\), \(W_i^K\), and \(W_i^V\) are learnable linear projections for each head, and \(W^O\) is another learnable linear projection for the output.

\subsubsection{Positional Encoding}

To handle positional information within input sequences, Transformer introduces positional encodings, which combine position information with word embeddings. The positional encoding can be mathematically defined as:

\begin{equation}
\text{PosEnc}(pos, 2i) = \sin\left(\frac{pos}{10000^{2i/d_{\text{model}}}}\right)
\end{equation}
\begin{equation}
\text{PosEnc}(pos, 2i+1) = \cos\left(\frac{pos}{10000^{2i/d_{\text{model}}}}\right)
\end{equation}

Here, \(pos\) represents the position of a token in the sequence, \(i\) is the dimension index, and \(d_{\text{model}}\) is the dimension of the model's hidden state.

\subsection{Unique Features of DeBERTa}

DeBERTa introduces the following innovative designs on top of BERT, making it suitable for the evaluation of clinical patient records:

\subsubsection{Disentangled Attention}

DeBERTa utilizes Disentangled Attention, a mechanism that decouples attention weights traditionally seen in the Transformer, to better differentiate information at different positions. The formula for Disentangled Attention is as follows:

\begin{equation}
\text{Attention}_{\text{Disentangled}}(Q, K, V) = \text{Softmax}\left(\frac{Q(K + b)^T}{\sqrt{d_k}}\right)V
\end{equation}

Here, \(Q\), \(K\), and \(V\) represent matrices for queries, keys, and values, \(d_k\) is the dimension of attention heads, and \(b\) is a learned bias vector. This mechanism calculates attention weights separately for different positions with the addition of bias \(b\), enabling the model to more effectively capture semantic relationships between different positions in the text.

\subsubsection{Decoding Enhancement}

DeBERTa introduces Decoding Enhancement, a mechanism that enhances the model during the decoding process, improving its performance in generative tasks

\subsection{Masked Language Modeling}

The implementation of MLM consists of the following key components:

\begin{enumerate}
    \item \textbf{Pretraining (MLM)}: Initially, the language model undergoes pretraining on a diverse corpus of text data, encompassing medical literature, clinical guidelines, and general medical texts.
    
    \item \textbf{Token Masking}: During MLM pretraining, a fraction of tokens within input sequences is randomly selected for masking, typically with a designated [MASK] token. The model's objective is to predict the original tokens behind these masks.
    
    \item \textbf{Training Objective}: MLM training revolves around maximizing the likelihood of predicting the original tokens given the masked input. This encourages the model to capture semantic relationships and language structure effectively.
    
    \item \textbf{Fine-Tuning for Domain Adaptation}: To adapt the pretrained language model to the clinical domain and the specific NBME competition task, fine-tuning is performed using the competition dataset. This step refines the model's representations for clinical context and language.
    
    \item \textbf{Evaluation Metrics}: Model performance is assessed using appropriate metrics, such as micro-average F1 score, to assess its accuracy in identifying clinical symptoms in patient notes.
\end{enumerate}

\subsection{Pseudo Labeling for Model Training}

Our model's training process leveraged pseudo labels to enhance learning from unlabeled data. Pseudo labeling involves generating labels for unannotated data based on a previously trained model's predictions, effectively using the model's own inference to augment the training dataset. 

\begin{figure}[h]
    \centering
    \includegraphics[width=0.48\textwidth]{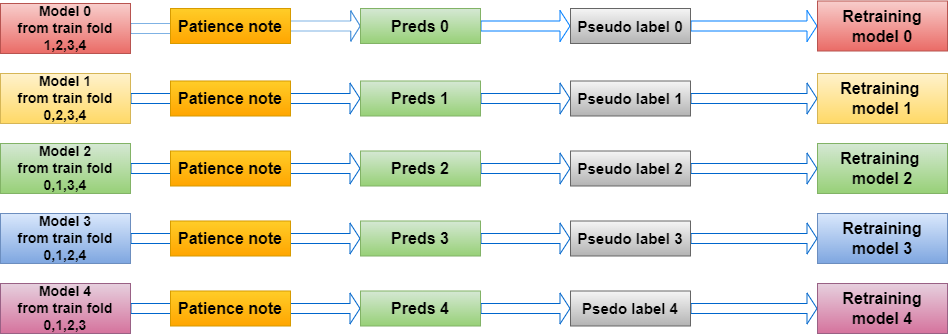}
    \caption{Schematic Diagram of Pseudo Label Generation.}
    \label{fig:Pseudo_Labels}
\end{figure}

The process commenced with initial model training using the standard dataset to learn fundamental patterns. Subsequently, pseudo labels were generated for unannotated data. Specifically, we employed the model to infer on a patient notes dataset containing 612,602 unlabeled IDs. This inference produced pseudo labels for the unannotated notes, as depicted in Figure \ref{fig:Pseudo_Labels}. The mathematical representation of pseudo label generation can be defined as:

\begin{equation}
\hat{y}_i = \text{argmax} \, P(y_i | x_i; \theta) 
\end{equation}

where \( \hat{y}_i \) represents the pseudo label for the i-th data point, \( x_i \) is the input, and \( P(y_i | x_i; \theta) \) is the probability of label \( y_i \) given input \( x_i \) as predicted by the model with parameters \( \theta \).

To prevent information leakage, especially concerning high similarity between pseudo labels and validation set texts, we directly generated pseudo labels instead of relying on averaged prediction results. This strategy ensured the integrity of our cross-validation scores.

In the subsequent stage, 50\% of the generated pseudo label data was selected for training. Training was limited to one epoch to allow the model to quickly assimilate the pseudo label data without overfitting. Fine-tuning on the original dataset followed to ensure the model's generalization capability.

Incorporating pseudo labels into our training regimen offered a dual advantage: it expanded the training data volume and introduced more variability and complexity, enabling the model to learn from a broader range of clinical scenarios. This approach significantly improved the model’s performance in clinical note interpretation, as evidenced by enhanced cross-validation scores.

\subsection{Optimization for Efficient Inference}

In typical NLP tasks with variable-length input sequences, padding is employed to ensure uniform sequence length for parallel processing by models like RNNs, CNNs, or Transformers.

\subsubsection{Standard Padding:}

Traditionally, sequences are padded to match the length of the longest sequence in the batch. Let \( L \) represent the maximum sequence length in a batch, and \( l_i \) be the length of the \( i \)-th sequence. The number of padding tokens added to a sequence is \( P_i = L - l_i \).

The computational cost for a batch \( B \) is given by:

\begin{equation}
\text{Cost}(B) = \sum_{i=1}^{n} P_i \cdot C
\end{equation}

where \( C \) is the computational cost of processing a single token, and \( n \) is the batch size.

\subsubsection{Padding Optimization Techniques:}

Optimization strategies include:

\begin{itemize}
\item Dynamic Batching: Grouping sequences with similar lengths to minimize padding.
\item Bucketing: Sorting sequences into buckets with similar lengths before batching.
\item Custom Padding Token Processing: Adapting the model to efficiently process padding tokens.
\end{itemize}

With these optimizations, the cost function changes. If \( L' \) is the optimized maximum sequence length, the new cost for a batch \( B' \) is:

\begin{equation}
\text{Cost}(B') = \sum_{i=1}^{n} P'_i \cdot C'
\end{equation}

where \( P'_i = L' - l_i \), and \( C' \) is the reduced computational cost per token.

\subsubsection{Impact on Inference Time:}

The impact on inference time \( T \) is given by:

\begin{equation}
T = \frac{\text{Cost}(B)}{\text{Cost}(B')}
\end{equation}

\begin{figure}[h]
    \centering
    \includegraphics[width=0.48\textwidth]{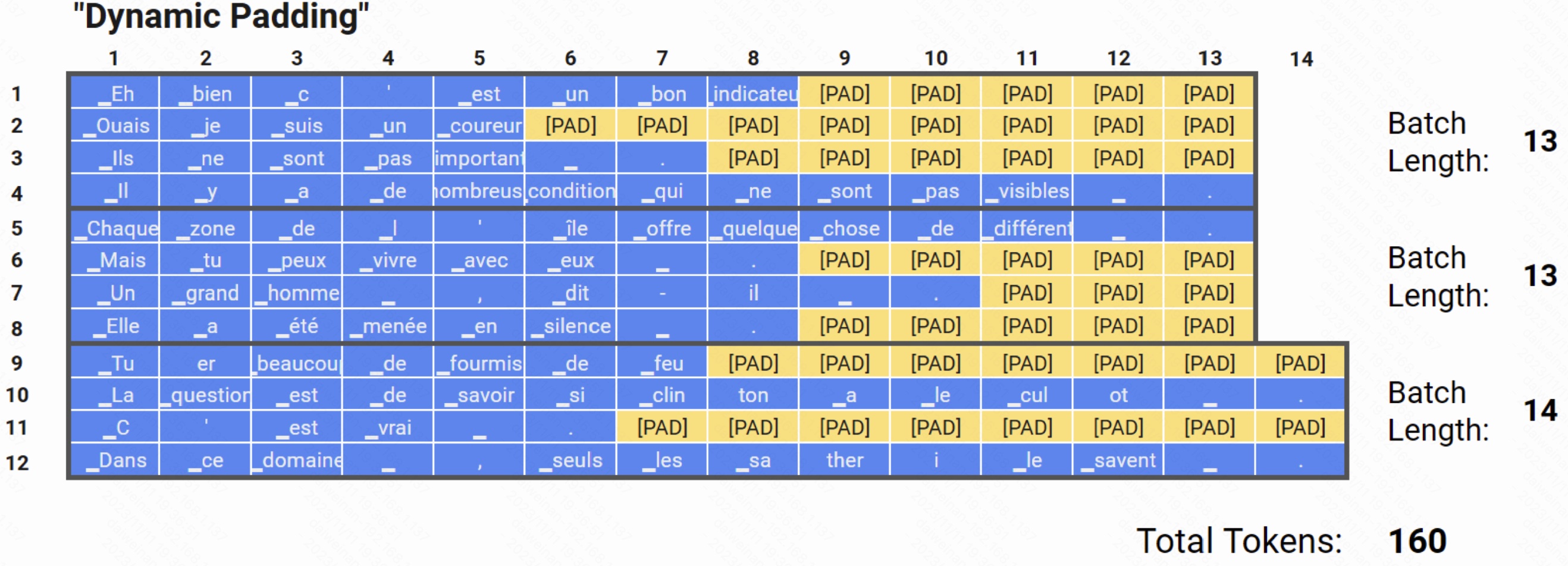}
    \caption{Fast Inference through Padding Optimization}
    \label{fig:fast_inference}
\end{figure}

A value of \( T < 1 \) indicates faster inference with optimized padding, resulting in performance gains, especially in large datasets or real-time applications. With this optimization, we reduced the inference time from 97 minutes to 56 minutes. The whole process is showned in Figure \ref{fig:fast_inference}.

\subsection{Loss Function}

We employ Binary Cross-Entropy with Logits Loss (BCEWithLogitsLoss) as our primary loss function. This loss function is ideal for binary classification tasks.

Mathematically, BCEWithLogitsLoss is defined as:

\begin{equation}
\text{Loss}(x, y) = - (y \cdot \log(\sigma(x)) + (1 - y) \cdot \log(1 - \sigma(x)))
\end{equation}

Where:
\begin{itemize}
\item \(x\) represents the model's raw logits.
\item \(y\) is the ground truth label.
\item \(\sigma(x)\) is the sigmoid activation function applied to the logits.
\end{itemize}

This loss function measures the alignment between predicted character spans and ground-truth spans in the text data. It plays a crucial role in optimizing the model's parameters during training, enabling precise character span predictions for clinical patient notes.

\subsection{Evaluation Metrics}

Our primary evaluation metric is the micro-averaged F1 score, which balances precision and recall and is computed as:

\begin{equation}
F1_{\mu} = 2 \cdot \frac{P_{\mu} \cdot R_{\mu}}{P_{\mu} + R_{\mu}}
\end{equation}

\section{Experiment Results}

\subsection{Comparison Experiment with DeBERTa-v3-large using MLM and Pseudo Labeling}
we conducted a comparison experiment using the DeBERTa-v3-large model with Masked Language Modeling (MLM) and Pseudo Labeling. The results are summarized in Table \ref{table:case1}.

\begin{table}[h]
\centering
\caption{DeBERTa-v3-large with MLM and Pseudo Labeling}
\begin{tabular}{|l|l|l|l|l|}
\hline
Modle & Baseline & MLM & Pseudo Label & CV Score \\ \hline
DeBERTa-v3-large & Y & - & - & 0.8879 \\
\hline
DeBERTa-v3-large & N & Y & - & 0.8901 \\
\hline
DeBERTa-v3-large & N & Y & Y & 0.8911 \\ \hline
\end{tabular}
\label{table:case1}
\end{table}

Table \ref{table:case1} shows the cross-validation (CV) scores for different experiments. The "Baseline" column indicates whether the model was trained on a baseline dataset, the "Pretrain (mask\_prob=0.15)" column indicates whether pretraining with a masking probability of 0.15 was performed, and the "Pseudo Label" column indicates whether pseudo labeling was employed. The "CV Score" column represents the cross-validation score achieved for each experiment.

\subsection{Experiment Results after MLM and Pseudo Label Training on Models}
we evaluated the performance of various models after MLM pretraining and pseudo label training on baseline models. The results are summarized in Table \ref{table:case2}.

\begin{table}[h]
\centering
\caption{Results for Case 2: Models after Pretraining and Pseudo Labeling}
\begin{tabular}{|l|l|}
\hline
Model Name & CV Score \\ \hline
deberta-v3-large & 0.8911 \\
\hline
deberta\_v2\_xlarge & 0.8886 \\
\hline
deberta\_v2\_xxlarge & 0.8904 \\ \hline
\end{tabular}
\label{table:case2}
\end{table}

Table \ref{table:case2} displays the CV scores for different models after pretraining and pseudo label training. These models include deberta-v3-large, deberta\_v2\_xlarge, and deberta\_v2\_xxlarge. The CV scores represent the model's performance on the evaluation dataset after these training procedures.

These experimental results demonstrate the impact of pretraining and pseudo labeling on the model's performance across different configurations.

    \section{Conclusion}
Our study presents a comprehensive approach to automating the evaluation of clinical patient notes, a critical task in medical education and certification. We leverage advanced Natural Language Processing (NLP) techniques, focusing on Masked Language Modeling (MLM) pretraining and utilizing the micro-averaged F1 score as our primary evaluation metric. Our methodology includes innovative strategies such as data preprocessing, pseudo-labeling, and training acceleration, resulting in significant enhancements in model performance. By optimizing loss functions and training strategies, we demonstrate the potential for efficient and accurate clinical patient note scoring, contributing to the automation of essential assessments in the medical field.

 \bibliographystyle{IEEEtran}
    \bibliography{references}

\end{document}